\documentclass[11pt,a4paper]{article}
\usepackage[hyperref]{naaclhlt2019}
\usepackage{times}
\usepackage{latexsym}
\usepackage{graphicx}
\usepackage{booktabs}
\usepackage{url}
\usepackage{amsmath}
\usepackage{mathtools}
\usepackage{algorithm2e}
\aclfinalcopy 

\title{Improving Dialogue State Tracking by Discerning the Relevant Context}

\author {Sanuj Sharma, Prafulla Kumar Choubey, Ruihong Huang \\
         Department of Computer Science and Engineering\\
		Texas A\&M University\\
         {\tt (sanuj, prafulla.choubey, huangrh)@tamu.edu}}
\date{}

\begin{document}
\maketitle
\begin{abstract}
A typical conversation comprises of multiple turns between participants where they go back-and-forth between different topics. At each user turn, dialogue state tracking (DST) aims to estimate user's goal by processing the current utterance. However, in many turns, users implicitly refer to the previous goal, necessitating the use of relevant dialogue history. Nonetheless, distinguishing relevant history is challenging and a popular method of using dialogue recency for that is inefficient.  We, therefore, propose a novel framework for DST that identifies relevant historical context by referring to the past utterances where a particular slot-value changes and uses that together with weighted system utterance to identify the relevant context. Specifically, we use the current user utterance and the most recent system utterance to determine the relevance of a system utterance. Empirical analyses show that our method improves joint goal accuracy by 2.75\% and 2.36\% on WoZ 2.0 and MultiWoZ 2.0 restaurant domain datasets respectively over the previous state-of-the-art GLAD model.
\end{abstract}

\section{Introduction}
\begin{figure}[t]
  \center
  \includegraphics[width=0.95\columnwidth]{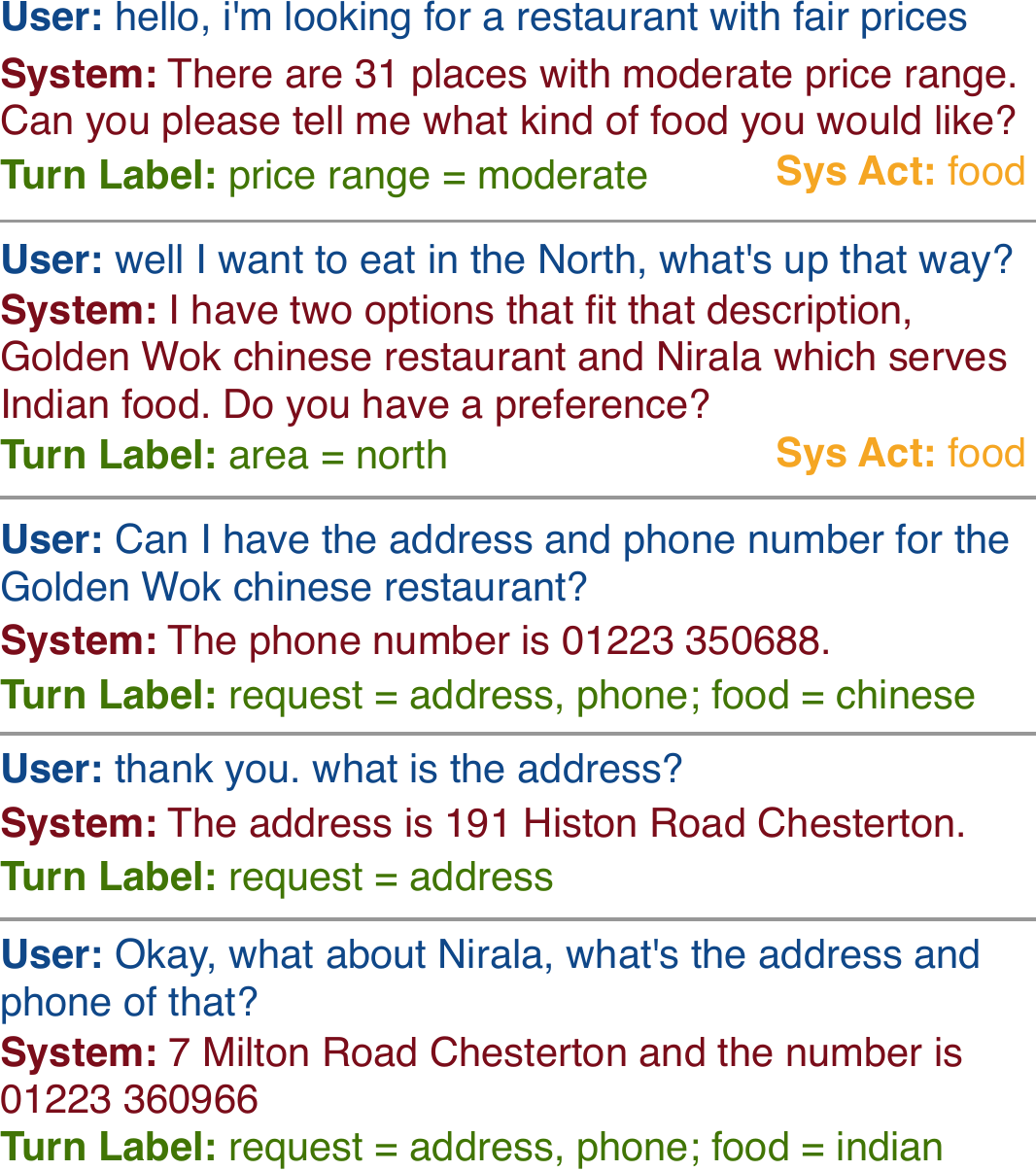}
  \caption{An example dialog from WoZ 2.0 dataset. A turn contains user utterance (blue), system utterance (red), system actions (yellow) and turn label (green). Each turn is separated by a line.}
  \label{fig:dialog}
\end{figure}

Dialog state tracking (DST) is a vital component in the task-oriented dialog systems which is used to estimate user's goals and requests in order to plan next action and respond accordingly. At each turn, DST aims to identify the set of goals that a user aims to achieve and requests that are represented as slot-value pairs. Typically, this decision is made by considering user utterance in the current turn or system actions in the previous turn. However, in many cases, the considered user utterance or system actions do not present enough information and refers to a previous utterance. 

As shown through an example in Figure \ref{fig:dialog}, while exploring different available options, user can go back-and-forth between the currently and previously discussed facts. For instance, when offered with two different restaurant options namely \textit{Nirala} \texttt{(food=indian)} and \textit{Golden Wok} \texttt{(food=chinese)} in the second turn, user first inquires about the details of {\it Golden Wok}. And after getting relevant details about the {\it Golden Wok} in the following two turns, user refers back to the second option provided in second turn and asks about {\it Nirala} restaurant. To predict the correct slot-value pair \texttt{food=indian} in the dialog state of the fifth turn, the system is required to refer back to the second turn again to find information about \textit{Nirala}, as the context obtained from the current dialog turn is insufficient. 

Identifying such implicitly referenced historical turns 
is challenging since implicit references are not local and most recent turns are often not informative. Therefore, the traditional approach of modeling dialogue recency \cite{W17-5526} may not suffice. 
Instead, we propose to  model implicit references by storing links to the past turn where each of the slots was modified. Then at each turn, we look up though the stored links to find the previous turn which may provide additional cues for predicting the appropriate slot-value.

Moreover,  the dialogue system often asks polar questions with yes-no answers.  For instance, the DST system should update the dialogue state with \texttt{food=indian} when a user replies {\it Yes} to a system utterance
{\it Do you want Indian food?}. 
In such cases, neither the user utterance nor system acts ({\it food} in this example) contain any information about the actual slot-value. 
This makes utilization of both system and user utterance eminent for dialog state tracking. However, 
utilizing the previous system utterance together with the current user utterance always at each turn may add noise.
Therefore, we use a gating mechanism based on both utterances 
to determine the relevance of the previous
system utterance in the current turn. 
The evaluation shows that identifying the relevant context is essential for dialogue state tracking. Our novel model that discerns important details in  non-adjacent dialogue turns and the previous system utterance from a dialog history is able to improve the previous state-of-the-art GLAD \cite{P18-1135} model on all evaluation metrics for both WoZ and MultiWoZ (restaurant) datasets. Furthermore, we empirically show that a simple self-attention based biLSTM model, using only one-third of the number of parameters as GLAD, outperforms GLAD by identifying and incorporating the relevant context.

\section{Related Work}

Early work for DST relied on separate Spoken Language Understanding (SLU) module \cite{henderson2012discriminative} to extract relevant information from user utterances in a pipelined approach. Such systems are prone to error accumulation from a separate SLU module, in absence of necessary dialog context required to interpret the user utterance. Thus, later work on DST moved away from separate SLU modules and inferred the dialog state directly from user utterance and dialog history \cite{W14-4340, henderson2014word, zilka2015incremental}. These models depend on delexicalization, using generic tags to replace specific slot types and values, and handcrafted semantic dictionaries. In practice, it is difficult to scale these models for every slot type and recent state-of-the-art models for DST use deep learning based methods to learn general representations for user and system utterances and previous system actions, and predict the turn state 
\cite{W13-4073, W14-4340, P15-2130, P17-1163, hori2016dialog, liu2017end, dernoncourt2017robust, chen2016end}. However, these systems are found to perform poorly on rare and unknown slot-value pairs which was recently addressed through local slot-specific encoders \cite{P18-1135} and pointer network \cite{P18-1134}.

A crucial limitation to all these approaches lies in the modeling of appropriate historical context, which is simply ignored in most of the works. Since user's goal may change back-and-forth between previous values, incorporating relevant historical context is useful in monitoring implicit goal references. In a recent work, \citet{W17-5526} discussed on similar limitations of current DST task and introduced a new task of frame tracking that explicitly tracks every slot-values that were introduced during the dialogue. However, that significantly complicates the task by maintaining multiple redundant frames that are often left unreferenced. Our proposed model, that explicitly track relevant historical user and system utterances, can be easily incorporated into any known DST or frame tracking systems such as  \citet{schulz2017frame} to replace the recency encoding.

\section{Discerning Relevant Context for DST}

\label{sec:model}
\begin{figure*}[t]
  \centering
  \includegraphics[width=0.95\textwidth]{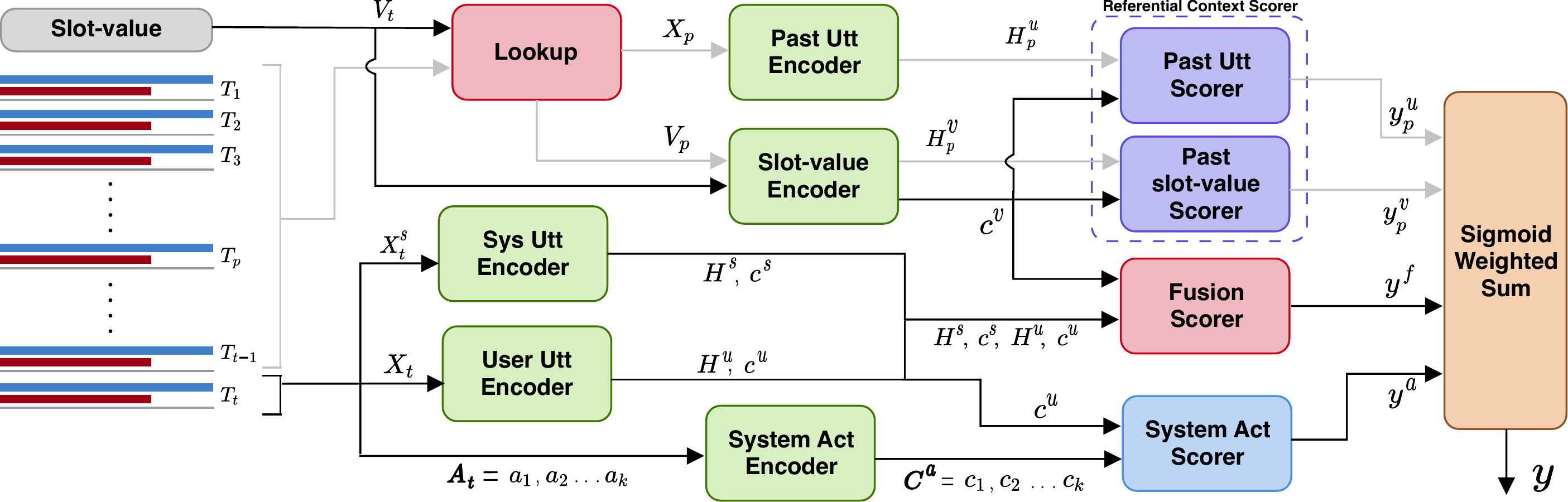}
  \caption{The Architecture of Context Aware Dialogue State Tracker.}
  \label{fig:model}
\end{figure*}

Similar to previous works, we decompose the multi-label classification problem to binary classification where we score each slot-value pair and select the ones that receive a score above a threshold to be included in the current dialog state. To predict the score for a candidate slot-value pair, the model uses 
the relevant past user utterance (referential utterance),
a fused utterance composed using the current user utterance and the system utterance of the previous turn, 
as well as previous system actions as evidence. 
Shown in Figure \ref{fig:model}, our model comprises of:




\vspace{.05in}
\noindent {\bf Lookup module:} retrieves a link to the turn where each of the slots changes. At each step, our system refers to the lookup module that returns the past 
user utterance (the ``{\it antecedent user utterance}'')
where the candidate slot-type was modified as well as outputs the 
previous slot-value. 

\vspace{.05in}
\noindent {\bf GLE modules:} Each of the five green modules in Figure \ref{fig:model} is a global-locally self-attentive encoder (GLE module) \cite{P18-1135} that encodes each type of evidence into a vector representation ($c$). Each input is represented as a sequence of words which is encoded to a vector representation via global-local self-attentive encoder (GLE) module \cite{P18-1135}. Specifically, GLE employs local slot-specific bidirectional LSTMs and a global bidirectional LSTM \cite{Hochreiter:1997:LSM:1246443.1246450} that is shared across all slots for encoding the input sequence into a sequence of hidden states ($H$), followed by a self-attention layer \cite{lin2016structured} to obtain a fixed dimension vector representation ($c$).

The GLE modules are used to encode the antecedent user utterance ($H^u_{p}, c^u_{p}$), the current user utterance ($H^u, c^u$), the previous system utterance ($H^s, c^s$), each of the system acts ($H^{a_i}, c^{a_i}$), as well as the previous slot-value ($H^v_{p}, c^v_{p}$) and the candidate ($H^v, c^v$) slot-value. 

\vspace{.05in}
\noindent {\bf Referential Context Scorer:} uses the candidate slot value ($c^v$),  the antecedent user utterance as well as the previous slot-value to 
determine if the candidate slot value was referenced  in the antecedent utterance.  
Specifically, the scorer uses the representation of the candidate slot value $c^v$ to attend over hidden states of the antecedent user utterance and the previous slot-value,  $H^u_{p}$ and $H^v_{p}$, and then computes attention weights for each of the hidden states. Next, the scorer sums up the hidden states weighed with the calculated attentions to get the summary context (Equation \ref{attn}). Finally, the scorer applies a linear neural layer to calculate the scores $y^v_{p}$ and $y^u_{p}$ representing the likelihoods that the candidate slot-value is different from the previous slot-value and the candidate slot-value was unreferenced in the antecedent utterance (Equation \ref{referential}).
  \begin{equation}
    \begin{split}
    Q(H,c): \  a_j = (H_j)^\top c \ ; \ p = softmax(a)
    \\
    Q(H,c) = \sum_i p_i H_i
\end{split}
\label{attn}
\end{equation}
\vspace{-.2in}
\begin{equation}
    \begin{split}
    y_p^u  = W_p^u \ Q(H_p^u, c^v) + b_p^u  
    \\
    y_p^v  = W_p^v \ Q(H_p^v, c^v) + b_p^v  
\end{split}
\label{referential}
\end{equation}

\noindent {\bf Fusion Scorer:} leverages necessary details in the previous system utterance to enrich the current user utterance. 
First, we use a gating mechanism based on $c^s$ and $c^u$ that determines the relevance of the previous system utterance in the current turn. We concatenate $c^s$ and $c^u$ and use a linear layer with sigmoid activation to calculate the score $\alpha$ (Equation \ref{alpha}). Then, we use attention from $c^v$ over $H^s$ and $H^u$ to calculate context summaries ($l^s, l^u$), and combine the summary vectors by taking their normalized weighted sum based on $\alpha$. 
We finally apply a single linear layer to calculate the score $y^f$ that determines the likelihood of the candidate slot-value based on both the current user utterance and the previous system utterance (Equation \ref{summ}).    
\begin{equation}
    \begin{split}
    f_c = W_{fc} (c^s \oplus c^u) + b_{fc} 
    \\
    \alpha = \sigma(W_{\alpha} tanh (f_c) + b_{\alpha})
\end{split}
\label{alpha}
\end{equation}
\vspace{-.2in}
\begin{equation}
    \begin{split}
    l^s = Q(H^s, c^v) \ ; \ l^u = Q(H^u, c^v)
    \\
    l^f = \alpha l^s + (1-\alpha)l^u \ ; \ y^f  = W_{lf} l^f + b_{lf}
\end{split}
\label{summ}
\end{equation}

\noindent {\bf System Act Scorer:} is the same as the action scorer proposed by \cite{P18-1135}. Specifically, The scorer uses attention from $c^u$ over $C^a$ to calculate action summary followed by a linear layer with sigmoid activation to calculate the score $y^a$ that determines the relevance of the candidate slot-value based on the previous system actions (Equation \ref{act}). 
\begin{equation}
    \begin{split}
     l^a = Q(C^a, c^u) \ ; \quad  y^a = (l^a)^\top c^v
\end{split}
\label{act}
\end{equation}
It then calculates the final score of the candidate slot-value by taking weighted sum of the four scores ($y_p^u$, $y_p^v$, $y^f$, $y^a$) followed by a sigmoid layer, where weights are learned in the network.

\section{Evaluations}

\subsection{Experimental Setup}

We primarily use WoZ 2.0 \cite{E17-1042} restaurant reservation task dataset that consists of 1200 dialogues for training and evaluation. Each dialogue has an average of eight turns, where each turn contains \textit{system utterance transcript}, \textit{user utterance transcript}, \textit{turn label} and \textit{belief state}.  All the dialogue states and actions are based on a task ontology that supports three different informable slot-types namely \textit{price range} with 4 values, \textit{food} with 72 values, \textit{area} with 7 values, and \textit{requests} of 7 different types like \textit{address} and \textit{phone}. Following the standard settings, we use 600 dialogues for training, 200 for validation and the remaining 400 for testing. 

We also use dialogues from restaurant domain in MultiWoZ 2.0 dataset \cite{budzianowski2018multiwoz} for secondary evaluation. It banks on a significantly complex ontology covering seven informable slot types with 276 different values (\textit{food, price range, restaurant name, area, book time, book day} and \textit{book people} with 97, 6, 105, 8, 43, 8 and 9 values respectively). We use standard training, validation and test splits of 1199, 50 and 61 dialogues respectively.




  


All the models on WoZ 2.0 are evaluated on the two standard metrics introduced in \citet{henderson2014third}. First, \textbf{Joint Goal Accuracy} is the percentage of turns in a dialogue where the user's informed joint goals are identified correctly. Joint goals are accumulated turn goals up to the current dialog turn. Second, \textbf{Turn Request Accuracy} calculates the percentage of turns in a dialogue where the user's requests were correctly identified. Models on MultiWoZ 2.0 dataset are evaluated using joint goal and turn inform accuracies, as used by \citet{nouri2018toward}. 

\subsection{Implementation Details}
We use pretrained GloVe word embeddings \cite{D14-1162} concatenated with character n-gram  embeddings \cite{D17-1206} which are kept fixed during the training. Each of bi-LSTMs use 200 hidden dimensions. All the models are trained using ADAM optimizer \cite{journals/corr/KingmaB14} with the initial learning rate of 0.001. Dropout rate  \cite{JMLR:v15:srivastava14a} is set to 0.2 for all biLSTM modules and the embedding layer. The models are trained for a maximum of 100 epochs with a batch size of 50. The validation data was used for early stopping and hyperparameter tuning.


\subsection{Results}

\begin{table}
\small
\centering
\resizebox{\columnwidth}{!}{
\begin{tabular}{ l c c }
\toprule
 & \multicolumn{2}{c}{WoZ 2.0} \\
 Model & Joint Goal & Turn Request \\
 \midrule
 Delexalisation-Based Model + SD & 83.7\% & 87.6\% \\
 NBT - DNN & 84.4\% & 91.2\% \\
 NBT - CNN & 84.2\% & 91.6\% \\ 
 GLAD $\dagger$ & {\bf 86.4\%} & {\bf 97.1\%} \\ \hline
 Global biLSTM based GLE & 85.0\% & 96.8\% \\
 Global biLSTM based GLE + RC & 87.4\% &  {\bf 97.0\%} \\
 Global biLSTM based GLE + RC + FS & \textbf{88.4\%} & \textbf{97.0\%} \\ \hline
 GLAD + RC + FS & \textbf{89.2\%} & \textbf{97.4\%} \\
 \bottomrule
\end{tabular}}
\caption{Test accuracy of baselines and proposed models on WoZ 2.0 restaurant reservation dataset. $\dagger$Retrained using docker container provided by the authors with exactly same hyper-parameters. We also experimented with different versions of PyTorch and cuDNN and found that results had high variance. 
Therefore, we report the average performance over 5 runs with different initializations for GLAD and all our models.}
\label{table:results}
\end{table}

\begin{table}
\small\addtolength{\tabcolsep}{10pt}
\resizebox{\columnwidth}{!}{
\begin{tabular}{ l c c }
\toprule
 & \multicolumn{2}{c}{MultiWoZ 2.0 (Restaurant)} \\
 Model & Joint Goal & Turn Inform \\
 \midrule
 GLAD & {43.95\%} & {76.99\%} \\
 GLAD + RC & \textbf{45.72\%} & \textbf{77.87\%} \\
 GLAD + RC + FS & \textbf{46.31\%} & \textbf{78.76\%} \\
 \bottomrule
\end{tabular}}
\caption{Test accuracy of GLAD and proposed models on MultiWoZ 2.0 restaurant domain dataset. Note that we considered all 276 slot-values for evaluating models. \citet{budzianowski2018multiwoz} reported joint goal accuracy of 80.9 on MultiWoZ 2.0 (restaurant) dataset. We believe they didn't include \textit{restaurant name} slot in their evaluation and only considered presence of three slot-types---\textit{book time, book day} and \textit{book people}---and not their values.}
\label{table:resultsmulti}
\end{table}

Table \ref{table:results} compares the performance of our proposed models with different baselines, including {\bf delexalisation-based model + SD} \cite{E17-1042}, DNN and CNN variants of {\bf neural belief tracker} \cite{P17-1163} and the previous state-of-the-art GLAD systems \cite{P18-1135} on WoZ 2.0 dataset. We also implement a simplified variant of GLAD, {\bf Global BiLSTM based GLE}, by removing slot-specific local biLSTMs from the GLE encoder. We then successively combine it with referential context ({\bf Global biLSTM based GLE + RC}) and the fused previous system utterance ({\bf Global biLSTM based GLE + RC + FS}). Finally, we directly incorporate the referential context and gate selected system utterance into the GLAD system ({\bf GLAD + RC + FS}).

Irrespective of the underlying system, utilizing appropriate context from the previous turns improves the overall performance of a dialogue state tracker on both joint goal and turn request accuracies on WoZ 2.0 dataset. First, incorporating relevant referential utterances to identify implicitly mentioned slot-value improves the accuracy of global biLSTM based GLE model on joint goal task by 2.4\%. Then, gating based mechanism to augment user utterance with relevant information from the previous system utterance further improves the joint goal accuracy by 1.0\%. Together, they improve joint goal and request accuracy of the  global biLSTM based GLE model by 3.4\% and 0.2\% respectively. Furthermore, as evident from the results in Table \ref{table:resultsmulti}, both referential context and fused system utterance proportionally improve performance on MultiWoZ 2.0 dataset as well with overall improvement of 2.36\% and 1.77\% on joint goal and turn inform accuracies respectively. Performances of all models on MultiWoZ 2.0 are significantly inferior compared to WoZ 2.0 owing to higher complexity, with richer and longer utterances and considerably more slot-values in the former dataset.




\section{Analysis}

\begin{table}
\centering
\resizebox{\columnwidth}{!}{
\begin{tabular}{ l c }
\toprule
 Model & Approx. \# of parameters \\
 \midrule
 Global biLSTM based GLE & 1.2 million \\  
 \textbf{Global biLSTM based GLE + RC + FS} & \textbf{6 million} \\
 GLAD & 17 million \\
 GLAD + RC + FS & 28 million \\
 \bottomrule
\end{tabular}}
\caption{Number of learnable parameters for different models on WoZ 2.0 dataset}
\label{table:parameters}
\end{table}


The utilization of relevant context results in significant reduction in the number of learnable parameters in the model as shown in Table \ref{table:parameters}. Relevant context with the baseline model is able to outperform GLAD 
while using only one third of the number of learnable parameters. The parameters added due to using relevant
context are the parameters for 
encoding the antecedent referential user utterance and the previous system utterance as well as the past utterance and past slot-value scorers. 
However, we also observe high variance in the joint goal accuracy. Since joint goal is calculated by accumulating turn goals, an error in predicting a turn goal is propagated to all the downstream turns. 


\section{Conclusion}
We have presented a 
novel method for identifying the relevant historical user utterance as well as determining the relevance of the  system utterance from the last turn to enrich the current user utterance and improve goal tracking in dialogue systems. 
The experimental results show that 
discerning relevant context 
from the dialog history is crucial for tracking dialog states.

 \section*{Acknowledgments}
We want to thank our anonymous reviewers for providing insightful review comments.


\bibliography{naaclhlt2019}
\bibliographystyle{acl_natbib}




\end{document}


\section{Encoders}

\begin{equation}
    H = BiLSTM(X) \quad \in \mathbb{R} ^ {n \times d}
\end{equation}

\begin{alignat}{4}
    &a_i \quad &&= \quad WH_i + b  &&\in  \mathbb{R} && \\
    &p   \quad &&= \quad softmax(a)  &&\in  \mathbb{R}^n && \\
    &c   \quad &&= \quad \sum_i p_i H_i  &&\in  \mathbb{R}^d &&
\end{alignat}

\begin{equation}
    H, c \quad = \quad Encode(X)
\end{equation}

\begin{alignat}{4}
    & H^u, c^u \quad &&= \quad Encode(U) && \\
    & H^s, c^s \quad &&= \quad Encode(S) && \\
    & H^a_i, C^a_i \quad &&= \quad Encode(A_i) && \\
    & H^v, c^v \quad &&= \quad Encode(V) && \\
    & H_p^u, c_p^u \quad &&= \quad Encode(U_p) && \\
    & H_p^v, c_p^v \quad &&= \quad Encode(V_p) &&
\end{alignat}

\section{Scorers}

\begin{alignat}{5}
    & a_j \quad &&= \quad (H_j)^\top c && \in \mathbb{R} && \\
    & p   \quad &&= \quad softmax(a) && \in \mathbb{R}^n && \\
    & l   \quad &&= \quad \sum_i p_i H_i && \in \mathbb{R}^d && \\
    & y   \quad &&= \quad Wl + b && \in \mathbb{R} &&
\end{alignat}

\begin{alignat}{4}
    & l \quad && = \quad attention(H, c) && \\
    & y \quad && = \quad Score(H, c) &&
\end{alignat}

\begin{alignat}{4}
    & y_p^u \quad &&= \quad Score(H_p^u, c^v) && \\
    & y_p^v \quad &&= \quad Score(H_p^v, c^v) &&
\end{alignat}

\subsection{Fusion Scorer}

\begin{alignat}{4}
    & l^s \quad &&= \quad attention(H^s, c^v) && \\
    & l^u \quad &&= \quad attention(H^u, c^v) && \\
    & fc  \quad &&= \quad W_1 (c^s \oplus c^u) + b_1 && \\
    & \alpha \quad &&= \quad \sigma(W_2 tanh (fc) + b_2) && \\
    & l^f    \quad &&= \quad \alpha l^s + (1-\alpha)l^u && \\
    & y^f    \quad &&= \quad W l^f + b &&
\end{alignat}

\subsection{Action Scorer}

\begin{alignat}{4}
    & l^a \quad &&= \quad attention(C^a, c^u) && \\
    & y^a    \quad &&= \quad (l^a)^\top c^v &&
\end{alignat}

\subsection{Analysis}